\documentclass[preprint,12pt]{elsarticle}

\usepackage[numbers,sort&compress]{natbib}
\usepackage{amssymb}
\usepackage{graphicx}
\usepackage{amsmath}
\usepackage{multirow}
\usepackage{caption}
\usepackage{tabularx}
\usepackage{geometry}
\usepackage{booktabs}
\usepackage{graphicx}
\usepackage{subcaption}
\usepackage{rotating}
\biboptions{numbers,sort&compress}

\newgeometry{left=2cm,right=2cm,top=2cm,bottom=2cm}

\journal{}
\begin{document}
\begin{frontmatter}
\title{FLOW: Fusing and Shuffling Global and Local Views for Cross-User Human Activity Recognition with IMUs}

\author[usc]{Qi Qiu}
\author[usc]{Tao Zhu}
\author[ust]{Furong Duan}
\author[akl]{Kevin I-Kai Wang}
\author[dut]{Liming Chen}
\author[usc]{Mingxing Nie}
\author[usc]{Yaping Wan}
\affiliation[usc]{organization={School of Computer Science, University of South China},
	city={Hengyang},
	postcode={430001}, 
	state={Hunan},
	country={China}}
	
\affiliation[ust]{
	organization={School of Computing, University of Ulster},
	postcode={BT37 0QB},
	city={Newtownabbey},
	country={Northern Ireland}}

\affiliation[akl]{
	organization={Department of Electrical, Computer, and Software Engineering, The University of Auckland},
	city={Auckland},
	postcode={1010},
	state={Auckland},
	country={New Zealand}}

\affiliation[dut]{
	organization={School of Computer Science and Technology, Dalian University of Technology},
	city={Dalian},
	postcode={116000},
	state={Liaoning},
	country={China}}








\begin{abstract}
Inertial Measurement Unit (IMU) sensors are widely employed for Human Activity Recognition (HAR) due to their portability, energy efficiency, and growing research interest. However, a significant challenge for IMU-HAR models is achieving robust generalization performance across diverse users. This limitation stems from substantial variations in data distribution among individual users.
One primary reason for this distribution disparity lies in the representation of IMU sensor data in the local coordinate system, which is susceptible to subtle user variations during IMU wearing.
To address this issue, we propose a novel approach that extracts a global view representation based on the characteristics of IMU data, effectively alleviating the data distribution discrepancies induced by wearing styles. To validate the efficacy of the global view representation, we fed both global and local view data into model for experiments. The results demonstrate that global view data significantly outperforms local view data in cross-user experiments.
Furthermore, we propose a Multi-view Supervised Network (MVFNet) based on Shuffling to effectively fuse local view and global view data. It supervises the feature extraction of each view through view division and view shuffling, so as to avoid the model ignoring important features as much as possible. Extensive experiments conducted on OPPORTUNITY and PAMAP2 datasets demonstrate that the proposed algorithm outperforms the current state-of-the-art methods in cross-user HAR.
\end{abstract}

\begin{graphicalabstract}
\end{graphicalabstract}

\begin{highlights}
\item Data-centric AI approach for cross-user HAR instead of transfer learning
\item Unifying coordinate representations to reduce data distribution differences
\item A novel traning method that shuffles and fuses global and local views
\item Finding that the global coordinate representation is more robust than local
\end{highlights}


\end{frontmatter}


\section{Introduction}
\label{introduction}
Human Activity Recognition based on Inertial Measurement Unit data (IMU-HAR), has been widely applied in fields such as health monitoring, smart care, and human-computer interaction due to its non-invasive and portable advantages \cite{lu2020driver, zhu2023diamondnet, wang2022sensor, chen2012sensor, wang2023negative}.

In recent years, many IMU-HAR schemes have been proposed \cite{ordonez2016deep, yuan2024self, tehrani2024wearable, qin2024improved}, DeepConvLSTM is a classic work that extracts features through a convolutional structure and adds a LSTM after the convolutional structure to better adapt to time-series classification tasks \cite{ordonez2016deep}. 
Wu et al. designed the Differential Spatio-Temporal LSTM (DST-LSTM) to learn dynamic spatial information, thereby improving activity recognition performance \cite{wu2023novel}.
Tang et al. constructed a multi-column neural network to learn bioinformatics from user data \cite{tang2022wmnn}.
Nouri-ani et al. presented an activity recognition system that utilizes a transfer function related to the corresponding activity \cite{nouriani2023system}.
Wang et al. proposed a multimodal information fusion framework for recognizing operating activities \cite{wang2023learning}.

These schemes can achieve good results when the training set and test set have similar distributions. However, in the production environment, the data of the training set and the test set often come from different users. Due to the distribution differences between user data, the model that performs well on the training set does not perform well on new users. This phenomenon is often referred to as the cross-user problem \cite{zhao2020local, ye2024cross}.


Most of existing studies generally consider that the cross-user problem arises from differences in data distribution across users. To address the challenge of limited model generalization due to data distribution shifts, researchers have explored various transfer learning techniques, such as optimal trasport \cite{ye2024deep},  generative adversarial network \cite{sanabria2021contrasgan} and de-entanglement scheme \cite{qian2021latent}. 

\begin{figure}[htbp]
	\centering
	\includegraphics[width=\linewidth]{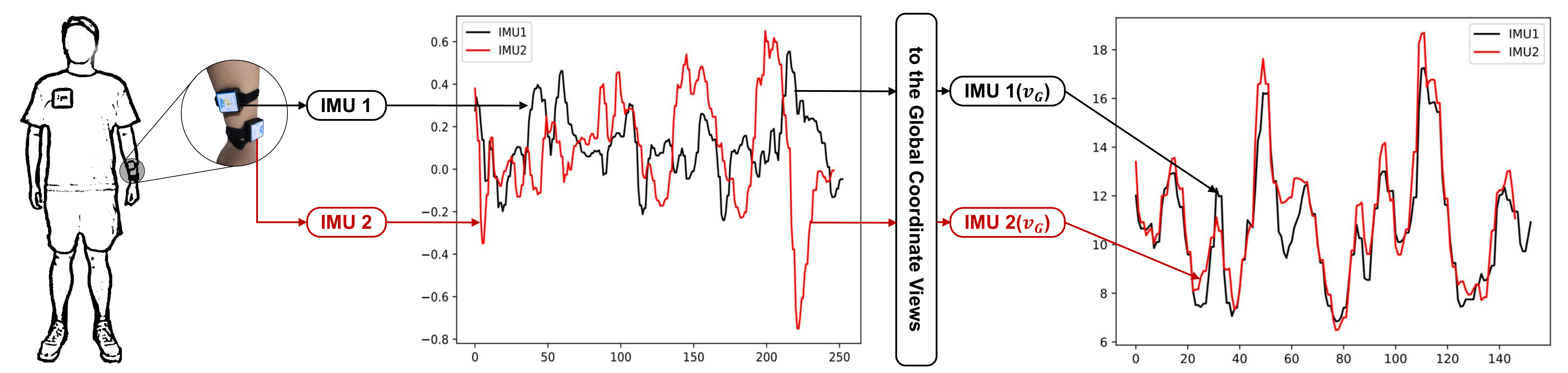}
	\caption{The NED coordinate representation can relieve the distribution differences caused by the wearing style. The left part of the image shows diagrams of two different IMU wearing styles. The line graph in the middle part shows the data of two IMU sensors in the local coordinate system, and the right part shows the data representation in the NED coordinate system. It can be seen that the data of the two IMUs is more closely represented in the latter.}
	\label{fig1}
\end{figure}

Different from these transfer learning method, this paper explores the specific reasons of data distribution differences among users based on the characteristics of IMU-HAR. 
We found that a key reason is the attitudes of IMUs are different. As shown in Fig~\ref{fig1}, two IMU sensors with different attitudes are worn on the same arm of the user. 
The middle part of Fig~\ref{fig1} shows the line graphs of the "z-axis acceleration" of the two sensors in the same time period. Although both describe the movement of the same arm and contain similar information, the information represented is very different due to the use of their own local coordinates as the basis.

The inherent discrepancies in IMU data arise from the sensors operating in distinct coordinate systems. By transforming both datasets into a unified spatial coordinate system, these inconsistencies can be effectively mitigated. To address this challenge, we propose a novel approach that employs the Mahony attitude solution algorithm and coordinate base transformation to convert IMU data from a local coordinate representation to a global one. The global coordinate system we adopted is the North-East-Down (NED) convention \cite{cai2011coordinate}. The line chart on the right of Fig~\ref{fig1} shows the line graphs of the "z-axis acceleration" of the two IMU data after conversion in the same time period. We call the data after base transformation as global views data ($\boldsymbol{v}_G$), and the data before transformation as local views data ($\boldsymbol{v}_L$). 
 

Transforming IMU data from the local coordinate system to the NED coordinate system can be divided into two parts: Mahony attitude solving algorithm  \cite{VASCONCELOS20088599} and changing of base, we refer to this process as the "M\&C" in the following description. It is worth noting that the M\&C process incurs critical information loss, and therefore a model trained solely with $\boldsymbol{v}_G$ may not necessarily outperform $\boldsymbol{v}_L$. In this paper, we propose to fuse $\boldsymbol{v}_G$ and $\boldsymbol{v}_L$ for HAR.

This approach aligns with the principles of data-centric AI \cite{zha2023data}, where the emphasis on data quality takes center stage in constructing robust AI systems. This paradigm shift prioritizes ensuring data quality and reliability over solely focusing on model improvement. Data-centric AI has demonstrably yielded benefits across diverse domains, including natural language processing \cite{chowdhary2020natural}, computer vision \cite{voulodimos2018deep}, and recommender systems  \cite{zhang2019deep}. Notably, there is a scarcity of research specifically applying a data-centric approach to enhance the cross-user performance of IMU-HAR models


Beyond distribution differences, model overfitting due to data scarcity can severely limit a model's ability to generalize to new user data. Consider an example: imagine a training dataset where all standing data is collected from users facing south, while all sitting data comes from users facing north. In this scenario, the model might exploit this  bias and solely rely on magnetometer data (sensitive to direction) to classify activities perfectly on the training set. However, this reliance on a single feature for differentiation makes it vulnerable to failing on unseen data where users perform these activities in different orientations.

Similar problems arise in the fusion of $\boldsymbol{v}_G$ and $\boldsymbol{v}_L$. If the model can achieve high training accuracy by solely relying on  $\boldsymbol{v}_L$ information, incorporating $\boldsymbol{v}_G$ becomes redundant. This not only renders the extraction of  $\boldsymbol{v}_G$ meaningless but also potentially limits the model's overall ability to capture rich information from the sensors.

In response to the problem of model over-reliance on certain features, this paper proposes a new training scheme called multi-view fusion training and its corresponding network architectures multi-view fusion net(MVFNet), which is used to ensure that as many features as possible participate in the final decision of the model. The basic idea is to shuffle and reassemble multiple samples by view, and supervise each view separately at the position near the output layer of the model to ensure that as many features as possible are used for the final decision. 

Because HAR tasks often require utilizing information from multiple sensors, and this information may manifest in different modalities, fusion operations are not uncommon in HAR tasks. However, the fusion mentioned here differs from previous fusion methods. Previous research on data fusion often aimed to unify information from different representations or allow information from different sources to complement each other, thereby improving model performance \cite{sun2022human}. 

In contrast, in this study, fusion primarily prevents redundant information that appears in the training set from being ignored by neural networks. This information may play an important role in the test set \cite{munzner2017cnn, kasnesis2019perceptionnet, bagadia2021fusion}. 
For example, $\boldsymbol{v}_G$ extracted from $\boldsymbol{v}_L$, and the $\boldsymbol{v}_G$ may be redundant when both are input to the network, but the representation generated from $\boldsymbol{v}_G$ could be valuable for the test set.

To verify the improvement effect of this paper in the IMU-HAR, we conducted experiments on the two most commonly used IMU datasets and achieved results exceeding the current latest scheme. In order to confirm the contribution of each part to the improvement of the effect, we conducted a number of ablation experiments.

In summary, the main contributions of this paper are as follows:

\begin{itemize}
	\item \textbf{Coordinate Transformation for Data Distribution Alignment:} This study proposes a method to transform IMU data from local coordinates to global coordinates (NED coordinates), effectively reducing the distribution difference caused by IMU attitude differences. Experiments demonstrate that models trained with global coordinates exhibit more robust features in cross-user HAR, even though coordinate transformation introduces some information loss.
	\item \textbf{Multi-view Fusion for Robust Feature Learning:} This paper presents a novel model named Multi-view Fusion Network (MVFNet) for fusing data from two different coordinate representations. Unlike previous methods that fuse different modalities to achieve information complementation, MVFNet randomly shuffles samples and independently supervises different views to encourage the model to utilize richer features for classification; meanwhile, MVFNet incorporates a novel learnable voting structure to integrate classification results based on different features, ultimately achieving more robust HAR.
	\item \textbf{Data-centric Approach for Cross-user HAR: } This study proposes a new approach to cross-user IMU-HAR that goes beyond transfer learning. The idea is data-centric, and changes the representation scheme of the data to reduce the distribution difference between different users' data, and fuses the new representation scheme with the original representation scheme to avoid information loss. Experimental results show that the proposed scheme performs more robust than the current latest transfer learning schemes on cross-user HAR, which proves the potential of the data-centric method in improving the generalization and robustness of the model in IMU-HAR.
\end{itemize}

The next part: In Section 2, we will briefly introduce the IMU-HAR definition and related domain generalization research of this study; in Section 3, we will introduce the attitude feature extraction scheme, the new feature fusion method, and the improved self-training method of this paper, as well as the complete cross-user problem solution; Section 4 is the experimental part, which will test the scheme proposed in this paper on the two most commonly used data sets in the IMU-HAR field, compare it with the current popular schemes, and conduct corresponding ablation experiments; Section 5 concludes the paper with future directions.

\section{Related Work}
\label{related_work}
\subsection{IMU-HAR}



Human activity recognition (HAR) has witnessed a surge in interest recently. Researchers are actively developing accurate and robust activity recognition models.

For instance, Mohamed et al. \cite{abdel2020st} proposed ST-deepHAR to leverage both temporal and spatial information. Guo et al. \cite{guo2021evolutionary} presented a dual-ensemble method that combines two ensemble models for enhanced activity classification. Building upon these advancements, Liu et al. \cite{liu2023distributional} introduced the Distributional and Spatial-Temporal Robust Representation (DSTRR), which learns statistical, spatial, and temporal features within a unified framework.

IMU-based HAR (IMU-HAR) can be categorized as a time series segment classification task. A time series segment, denoted by $S$, can be represented as a matrix of dimensions $t \times c$, where $t$ represents the segment length and $c$ represents the feature vector length for each time step, also known as the number of channels.

Specifically, in IMU-HAR, $c$ is typically 9, encompassing data from three-axis accelerometers, gyroscopes, and magnetometers.

$$
S = 
\begin{pmatrix}
	a_x^1 & a_y^1 & a_z^1 & m_x^1 & m_y^1 & m_z^1 & g_x^1 & g_y^1 & g_z^1 \\
	a_x^2 & a_y^2 & a_z^2 & m_x^2 & m_y^2 & m_z^2 & g_x^2 & g_y^2 & g_z^2 \\
	\vdots & \vdots & \vdots & \vdots & \vdots & \vdots & \vdots & \vdots & \vdots \\
	a_x^t & a_y^t & a_z^t & m_x^t & m_y^t & m_z^t & g_x^t & g_y^t & g_z^t
\end{pmatrix}
$$

The core objective of IMU-HAR is to train a model, denoted by $M(\boldsymbol{x}; \boldsymbol{\theta})$, that can map a training dataset $X$ containing $N$ time series segments ($S^1, S^2, ..., S^N$) to a set of discrete class labels ($l \in \{1, 2, 3, ..., k\}$). This mapping is achieved while minimizing the cross-entropy loss function, defined as:

\begin{align} \label{Lce}
	\mathcal{L}_{\text{CE}} = -\dfrac{1}{N} \sum_{i=1}^N \sum_{j=1}^k y^i_j \log(\hat{y}^i_j)
\end{align}

where $\boldsymbol{y}^i$ represents the one-hot encoded label vector for segment $S^i$, and $\boldsymbol{\hat{y}}^i$ denotes the model's output for the same segment.

\subsection{Cross-User HAR}

Cross-user HAR aims to identify activities from IMU sensor data of a new user without including their data in the training set. Individual variations in body structure, activity patterns, and sensor wearing styles lead to distribution shifts between the source data distribution, $P(Y | S, \mathcal{D}_s) = \{(S_s^i, y_s^i)\}^N_s$, in the training set and the target user's data distribution, $P(Y | S, \mathcal{D}_t) = \{(S_t^i, y_t^i)\}^N_t$, where $\mathcal{D}_s \cap \mathcal{D}_t = \emptyset$. These distribution discrepancies cause the model, $M$, trained on the source data, $\mathcal{D}_s$, to underperform on the target user's data, $\mathcal{D}_t$.

Domain generalization techniques address this cross-user challenge by training a model, $M$, on the source data, $\mathcal{D}_s$, to minimize the expected risk, Eq~\ref{Lce}, on unseen target data distributions like $\mathcal{D}_t$. Here, $\boldsymbol{y}^i$ represents the one-hot encoded label for the target user's sensor data, $S^i_t$, and $\boldsymbol{\hat{y}}^i$ denotes the predicted output of the model, $M(S^i_t, \theta)$.

Cross-user approaches can be broadly classified into two categories: domain generalization and domain adaptation.Domain generalization relies exclusively on source domain data during model training. In contrast, domain adaptation methods leverage unlabeled or few labeled target domain data to enhance model performance on the target domain.The proposed method falls under the domain generalization category. Several classical domain generalization methods have been proposed in the literature.

DANN (Domain-Adversarial Neural Network Training) is a cross-domain learning method that leverages adversarial training. The DANN model comprises three components: a feature extractor, a label predictor, and a domain discriminator. The feature extractor learns data representations, the label predictor classifies the data, and the domain discriminator identifies the data's origin domain. Through adversarial training, DANN learns robust feature representations that generalize well to different domains \cite{ganin2016domain}.

JAN (Joint Adversarial Network) is another cross-domain learning method based on adversarial training. Similar to DANN, JAN employs a three-component architecture with a shared feature extractor between the discriminator and classifier. This shared architecture encourages both components to learn more robust feature representations \cite{long2017deep}.

Cross-user identification is a critical application in cross-domain learning, where numerous domain-specific methods have been proposed to address this challenge.

\begin{itemize}
	\item A person re-identification model using multiple wearable sensors \cite{chen2022salience} reduces data discrepancies between different sensors by aligning local and global sensor features.
	\item A generalizable person activity recognition method \cite{qian2021latent} decomposes activity data into separate person identification and activity category features.
	\item The Transformer-based TASKED approach leverages adversarial learning and MMD regularization to align data across domains, while employing self-supervised knowledge distillation to enhance model training robustness \cite{suh2023tasked}.
	\item An adversarial learning-based method improves the generalization ability of existing cross-user identification models \cite{leite2020improving}.
	\item DIN, a novel method, proposes a dual-branch network structure that combines the strengths of CNNs and Transformers. Its dual-stream architecture separates domain-specific and shared features, further enhancing model generalization \cite{tang2022dual}. 
\end{itemize}

We will compare our proposed method with these aforementioned approaches in the experiment section.

\section{Method}
\label{method}
This section details the proposed solutions to address cross-user variability in IMU-HAR.

In the first part, Global View Extraction, aims to mitigate the influence of IMU attitude on data distribution,  thereby reducing the distribution difference between users. This subsection delves into the process of extracting IMU attitude using the Mahony algorithm and subsequently obtaining a transformation matrix to convert local views ($\boldsymbol{v}_L$) to the global view ($\boldsymbol{v}_G$).

In the second part, Multi-view Fusion Training, we propose a scheme to effectively fuse $\boldsymbol{v}_L$ and $\boldsymbol{v}_G$. This part introduces the definition of view, the multi-view supervised model, and its training strategy.

Finally, this section presents the complete framework, named \textbf{F}using and Shuffling G\textbf{l}obal and L\textbf{o}cal Vie\textbf{w}s (FLOW), which integrates both the M\&C process and Multi-View Fusion Training. FLOW effectively addresses cross-user variability in IMU-HAR by mitigating sensor orientation effects and utilizing complementary information from local and global views.

\subsection{Global View Extraction By M\&C} \label{mandc}
The raw data from IMU sensors is typically represented in local coordinates. However, as described in the introduction, this representation can lead to discrepancies in data distribution among users. If the local coordinate representation can be unified under an objective global coordinate system, this will alleviate this problem.

\begin{figure}[htbp]
	\centering
	\includegraphics[width=\linewidth]{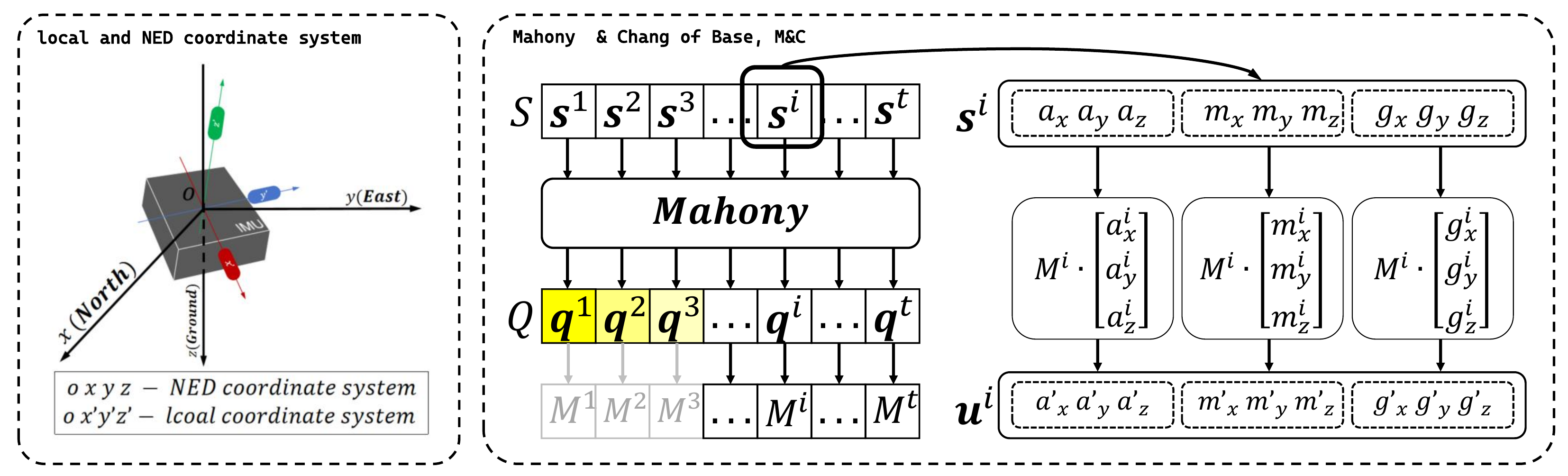}
	\caption{Two coordinate schemata and pose extraction schemes. The figure on the left shows the schematic diagrams of NED coordinate representation and local coordinate representation. The figure on the right shows the specific M\&C process. The original data is composed of a time series $S$ that is input into the Mahony algorithm to obtain the attitude representation $\boldsymbol{q^i}$ of the IMU for each moment. Then, the transformation matrix $M^i$ is obtained based on $\boldsymbol{q^i}$, and then the accelerometer, gyroscope, and magnetometer data in $\boldsymbol{s^i}$ are transformed from the local coordinate system to the NED coordinate system using $M^i$. However, due to the inherent characteristics of the Mahony algorithm, the attitudes in the first part of the obtained attitude sequence Q are inaccurate, and we will not use them in the experiment, which will result in the loss of about 1 second of data.}
	\label{fig2}
\end{figure}

The purpose of this subsection is to unify the representation of sensor data from different local coordinates to a representation in the world coordinate system, in order to relieve the influence of sensor attitude on data distribution. This process can be divided into two parts: finding the base coordinate transformation matrix using the Mahony and Changing (M\&C) method, as shown in Fig~\ref{fig2}

First, by employing the Mahony algorithm, we can transform the IMU local coordinates into the NED coordinate system. The NED coordinate system has the IMU center of mass as the origin, the true north direction as the x-axis direction, the true east direction as the y-axis direction, and the gravity direction as the z-axis direction. NED is a coordinate system that is independent of IMU attitude  as the right of Fig~\ref{fig2}.

This paper employs the Mahony attitude solution algorithm to obtain the sensor's attitude representation $\boldsymbol{q}$ in the NED coordinate system. As the Mahony algorithm is not the primary focus of this paper, only its input and output are presented without delving into its intricate details:

The Mahony algorithm takes as input a sequence of IMU nine-axis data points $S = [\boldsymbol{s}^1, \boldsymbol{s}^2, ..., \boldsymbol{s}^i, ..., \boldsymbol{s}^t]$, where each nine-axis data point $\boldsymbol{s}^i = [a_x, a_y, a_z, m_x, m_y, m_z, g_x, g_y, g_z]$  represents a specific time instant. 
It outputs a list of attitudes $Q = [\boldsymbol{q}^1, \boldsymbol{q}^2, ..., \boldsymbol{q}^i, ..., \boldsymbol{q}^t]$, where each quaternion $\boldsymbol{q}^i = [q_w, q_x, q_y, q_z]$ represents the IMU attitude at time $i$, defining the sensor's spatial orientation in the NED coordinate system. Consequently, the base coordinate transformation matrix from the local coordinate system to the NED coordinate system can be derived:

$$
M = \begin{bmatrix}
	1-2q_y^2-2q_z^2 & 2q_xq_y-2q_wq_z & 2q_xq_z+2q_wq_y \\
	2q_xq_y+2q_wq_z & 1-2q_x^2-2q_z^2 & 2q_yq_z-2q_wq_x \\
	2q_xq_z-2q_wq_y & 2q_yq_z+2q_wq_x & 1-2q_x^2-2q_y^2
\end{bmatrix} 
$$

Then, based on the transformation matrix, the coordinate representation in the gobal view is obtained:
\begin{align*}
	\boldsymbol{a'} &= M \cdot \begin{bmatrix} a_x, a_y, a_z \end{bmatrix}^T \\
	\boldsymbol{m'} &= M \cdot \begin{bmatrix} m_x, m_y, m_z \end{bmatrix}^T \\
	\boldsymbol{g'} &= M \cdot \begin{bmatrix} g_x, g_y, g_z \end{bmatrix}^T
\end{align*}

In this paper, we refer to $\boldsymbol{s}^i$ as the Local View($\boldsymbol{v}_L$) and $\boldsymbol{u}^i + \boldsymbol{q}^i = [a'_x, a'_y, a'_z, m'_x, m'_y, m'_z, g'_x, g'_y, g'_z, q_w$, $ q_x, q_y ,q_z]$ is referred to as the Global View ($\boldsymbol{v}_G$) of $\boldsymbol{s}^i$ [Fig~\ref{fig2}]. 

It is important to note that due to the random initial pose of the Mahony algorithm, the corresponding pose outputs for the first few time points might not be accurate. Consequently, the input time series required by the M\&C method needs to be sufficiently long and continuous. While this is not a limitation in real-world applications, it hinders the use of M\&C with pre-segmented datasets in experimental settings.

\subsection{Multi-View Fusion}

To fuse $\boldsymbol{v}_G$ and $\boldsymbol{v}_L$, we propose a multi-view fusion training scheme. This subsection will delve into the details of this multi-view supervised training method, encompassing view division, view-based shuffling, the multi-view fusion network, and its training process.

\subsubsection{Views Division} \label{secvd}

This subsection explores the concept of views and delves into the selection of view granularity.

First,we introduce the view definition. A data point, denoted by $\boldsymbol{x}^i$, can be interpreted as a combination of $n$ semantically complete parts, each referred to as a "view". In the context of IMU data, sensor categories can be used to divide the data into distinct views. For instance, acceleration, angular velocity, and magnetometer readings represent three separate views. Conversely, for multi-IMU activity recognition tasks, a coarser granularity can be adopted by treating the entire nine-axis data from a single IMU as a single view. In this scenario, the number of views would correspond to the number of IMU sensors involved in the task.

In this paper, we propose a multi-view fusion approach to effectively integrate $\boldsymbol{v}_L$ and $\boldsymbol{v}_G$. Therefore, for the remainder of this paper, we assume that the data from any IMU is divided into two views:

\begin{itemize}
	\item $\boldsymbol{v}_L$ \textbf{(view 1) :} The raw data $\boldsymbol{s}^i$ capture the fundamental motion information acquired by the IMU.
	
	\item $\boldsymbol{v}_G$ \textbf{(view 2) :} The representation $\boldsymbol{u}^i + \boldsymbol{q}^i$ in global coordinates obtained using the scheme in Section~\ref{mandc}.
\end{itemize}

Consequently, in a HAR task involving $m$ IMU sensors, its input will be divided into $2m$ views. 

The granularity of view division can influence the final performance of FLOW. In the ablation experiment section of this paper, we will delve into the advantages and disadvantages of different view division granularities. Additionally, for specific experiments, we will conduct evaluations based on the optimal granularity for their respective datasets. Therefore, the final partitioning strategy may differ from the one presented here.

\subsubsection{Shuffle}

During model training, we employ view-shuffled batching to ensure the model comprehends each view independently. This technique involves shuffling the data across views within a batch. In other words, each view within a sample can originate from a sample with a distinct label. Consequently, the model is tasked with predicting the label for each view in the shuffled sample. This enforces the model to learn the relationship between each individual view and the label, rather than solely focusing on the connection between the entire sample and the label.

Here's an illustrative example for a small batch training scenario. Consider a batch $X = \{(\boldsymbol{x}^i, y^i)\}^b$ of size $b$, where each sample $\boldsymbol{x}^i$ can be decomposed into $n$ views $[\boldsymbol{v}_1^i, \boldsymbol{v}_2^i, \dots, \boldsymbol{v}_n^i]$. We introduce a random $b \times n$ matrix $R$:

$$
\boldsymbol{R} = \begin{pmatrix}
	r_{11} & r_{12} & \cdots & r_{1n} \\
	r_{21} & r_{22} & \cdots & r_{2n} \\
	\vdots & \vdots & \ddots & \vdots \\
	r_{b1} & r_{b2} & \cdots & r_{bn} \\
\end{pmatrix} \quad (0 < r_{ij} < n )\quad \text{and} \quad (r_{ik} \neq r_{jk} \quad if \quad i \neq j)
$$

The view-based shuffle operation proceeds as follows for the generated samples $\tilde{\boldsymbol{x}}^i$ and their corresponding label $\tilde{\boldsymbol{y}}^i$:
\begin{align*}
	\tilde{\boldsymbol{x}}^i = [v_1^{r_{i1}}, v_2^{r_{i2}}, \dots, v_n^{r_{in}}] \\
	\tilde{\boldsymbol{y}}^i = [y^{r_{i1}}, y^{r_{i2}}, \dots, y^{r_{in}}]
\end{align*}

The resulting shuffled batch $\tilde{X}=\{(\tilde{\boldsymbol{x}}^i, \tilde{\boldsymbol{y}}^i)\}^b$ is then utilized for training in the current batch.


To illustrate the Shuffle process more clearly, let's consider a concrete example using a batch of size 4 $X = \{\boldsymbol{x}^1, \boldsymbol{x}^2, \boldsymbol{x}^3, \boldsymbol{x}^4\}$. Each sample $\boldsymbol{x}^i$ in this batch can be decomposed into three distinct viewpoints, represented as $\boldsymbol{x}^i = [v_1^i, v_2^i, v_3^i]$.

The random matrix $R$ is generated to control the shuffle process. 
Where each column of $R$ is a random permutation of the index of $X$, which is used to guide how to shuffle and reorganize the views.
$$
\boldsymbol{R} = \begin{pmatrix}
	3 & 4 & 1 \\
	1 & 3 & 2 \\
	2 & 1 & 3 \\
	4 & 2 & 4 \\
\end{pmatrix}\quad
$$

After applying the shuffle operation, the resulting batch $\tilde {X} = \{(\tilde {\boldsymbol{x}}^i, \tilde{\boldsymbol{y}}^i)\}^b$ is as follows: 
\begin{align*}
	\tilde{\boldsymbol{x}}^1 &= [v_1^3, v_2^4, v_2^1] & \tilde{\boldsymbol{y}}^1 &= [y^3, y^4, y^1] \\
	\tilde{\boldsymbol{x}}^2 &= [v_1^1, v_2^3, v_2^2] & \tilde{\boldsymbol{y}}^2 &= [y^1, y^3, y^2] \\
	\tilde{\boldsymbol{x}}^3 &= [v_1^2, v_2^1, v_2^3] & \tilde{\boldsymbol{y}}^3 &= [y^2, y^1, y^3] \\
	\tilde{\boldsymbol{x}}^4 &= [v_1^4, v_2^2, v_2^4] & \tilde{\boldsymbol{y}}^4 &= [y^4, y^2, y^4]
\end{align*}

\subsubsection{Multi-view Fusion Net and Training}
\begin{figure}[htbp]
	\centering
	\includegraphics[width=\linewidth]{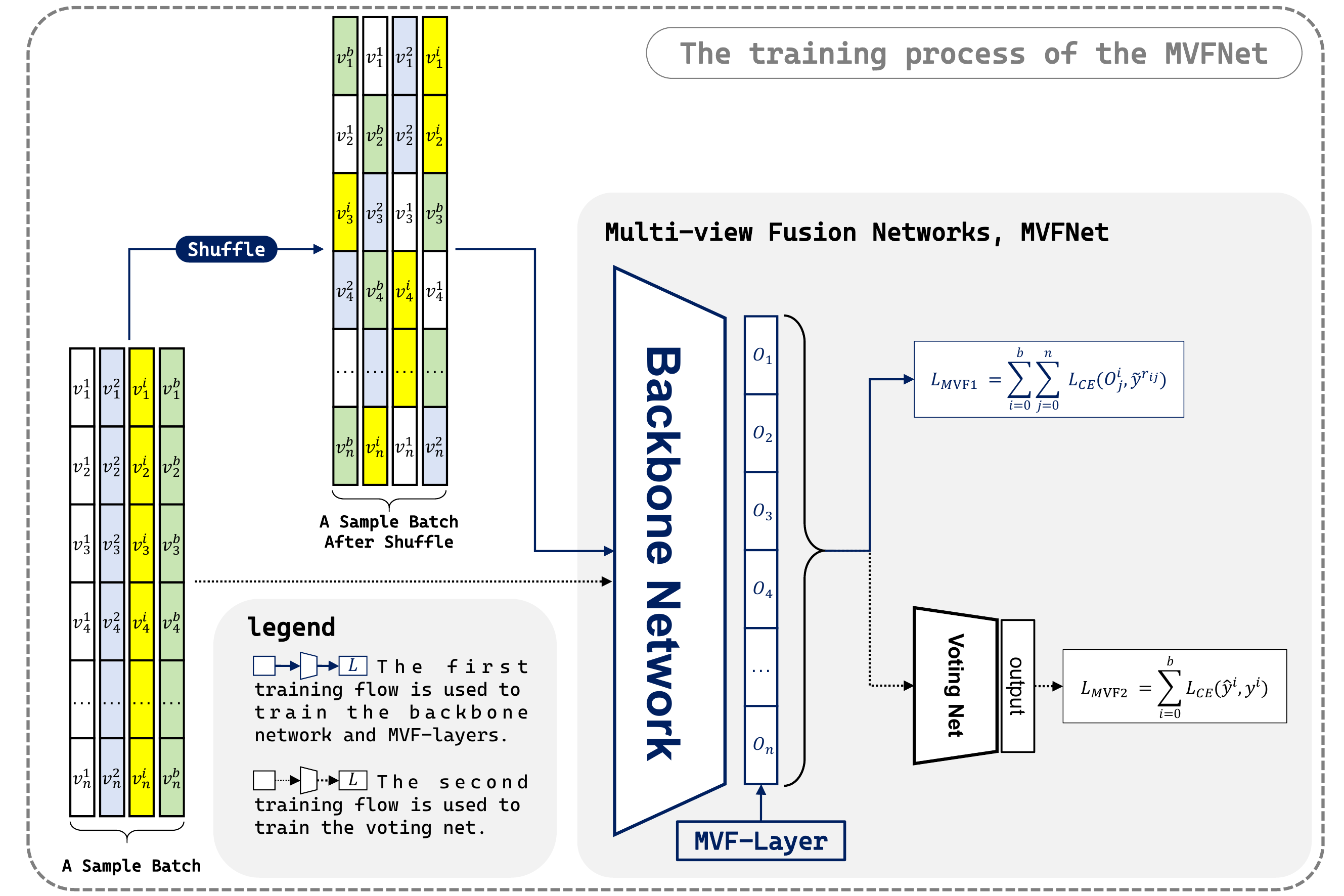}
	\caption{MVFNet and its training process. MVFNet consists of three parts: Backbone, MVF-Layer, and Voting Net. For the same batch of data, the training of MVFNet includes two different processes. In the first process, the data in the current batch is shuffled based on perspective and then used to train the Backbone and MVF-Layer. In the second process, we use the unshuffled data to train the Voting-Net with the Backbone and MVF-Layer frozen.}
	\label{fig3}
\end{figure}

Considering a HAR dataset where each sample $\boldsymbol{x}^i$ can be decomposed into $n$ views with complete semantics, then the corresponding multi-view supervised net(MVFNet) is shown in the Fig~\ref{fig3}. It consists of three parts namely the backbone network, a multi-view fusion layer (MVF-Layer) and a voting network (Voting-net).

The backbone network utilizes DeepConvLSTM, a well-established classification network in the HAR domain. while the MVF-layer, a fully connected layer, outputs a feature vector of dimension $k \times n$. This vector is then divided into $n$ groups, each containing $k$ elements corresponding to the $n$ views. Here, $k$ represents the number of categories in the classification task.
The Voting-net is a three-layer fully connected network, leverages the output from each group in the MVF-Layer to identify the view with the highest confidence score. During training, a batch of shuffled data $\tilde{X} = \{(\tilde{\boldsymbol{x}}^i, \tilde{\boldsymbol{y}}^i)\}^b$ is fed into the network. The corresponding loss function for training is defined as:
\begin{align}
	\mathcal{L}_{MVF1} = \sum_{i=1}^{b}\sum_{j=1}^{n} \mathcal{L}_{CE}(O^i_j, \tilde{y}^{r_{ij}})
\end{align}
where $\mathcal{L}_{CE}$ as Eq~\ref{Lce} represents the cross-entropy loss function and $O_j^i$ represents the output of the $j^{th}$ group of the MVF-layer for sample $\tilde{\boldsymbol{x}}^i$. 

Obviously, it can be seen from $\mathcal{L}_{MVF1}$ that the Voting-net is not trained, because there is no label that can describe a shuffled sample as a whole.
In fact, for a batch, MVFNet will go through two training procedures. The first training procedure is as described above. In the second training procedure, the data of the batch will not be shuffled, and the backbone network and MVF-layer will be frozen. Backpropagation is performed using the following loss function to train the Voting-net.
\begin{align}
	\mathcal{L}_{MVF2} = \sum_{i=0}^{b} \mathcal{L}_{CE}(\hat{y}^i_j, y^{i})
\end{align}
the $\hat{y}^i$ is the outputs of Voting-net given $\boldsymbol{x}^i$ as input.

The overall training process is shown in Fig~\ref{fig3}.

\subsection{FLOW}
This subsection combines "global view extraction" and "multi-view fusion training" to achieve IMU-HAR. 

Specifically, we use the original data $\boldsymbol{v}_L$ and the data after base transformation $\boldsymbol{v}_G$ as two different views for multi-view supervised training to achieve the fusion of the $\boldsymbol{v}_L$ and the $\boldsymbol{v}_G$. The overall scheme is called FLOW (\textbf{F}using and Shuffling G\textbf{l}obal and L\textbf{o}cal Vie\textbf{w}s).

Assume that in an activity recognition task, the number of IMU sensors worn by the user is $m$, and the model input for a time segment is the matrix $X$:

$$
X = 
\begin{pmatrix} 
	\boldsymbol{s}_1^1 & \boldsymbol{u}_1^1 & \boldsymbol{s}_2^1 & \boldsymbol{u}_2^1 & \dots & \boldsymbol{s}_m^1 & \boldsymbol{u}_m^1 \\ 
	\boldsymbol{s}_1^2 & \boldsymbol{u}_1^2 & \boldsymbol{s}_2^2 & \boldsymbol{u}_2^2 & \dots & \boldsymbol{s}_m^2 & \boldsymbol{u}_m^2 \\ 
	\vdots & \vdots & \vdots & \vdots & \vdots & \vdots & \vdots \\ 
	\boldsymbol{s}_1^t & \boldsymbol{u}_1^t & \boldsymbol{s}_2^t & \boldsymbol{u}_2^t & \dots & \boldsymbol{s}_m^t & \boldsymbol{u}_m^t \\
\end{pmatrix}
$$

\begin{figure}[htbp]
	\centering
	\includegraphics[width=\linewidth]{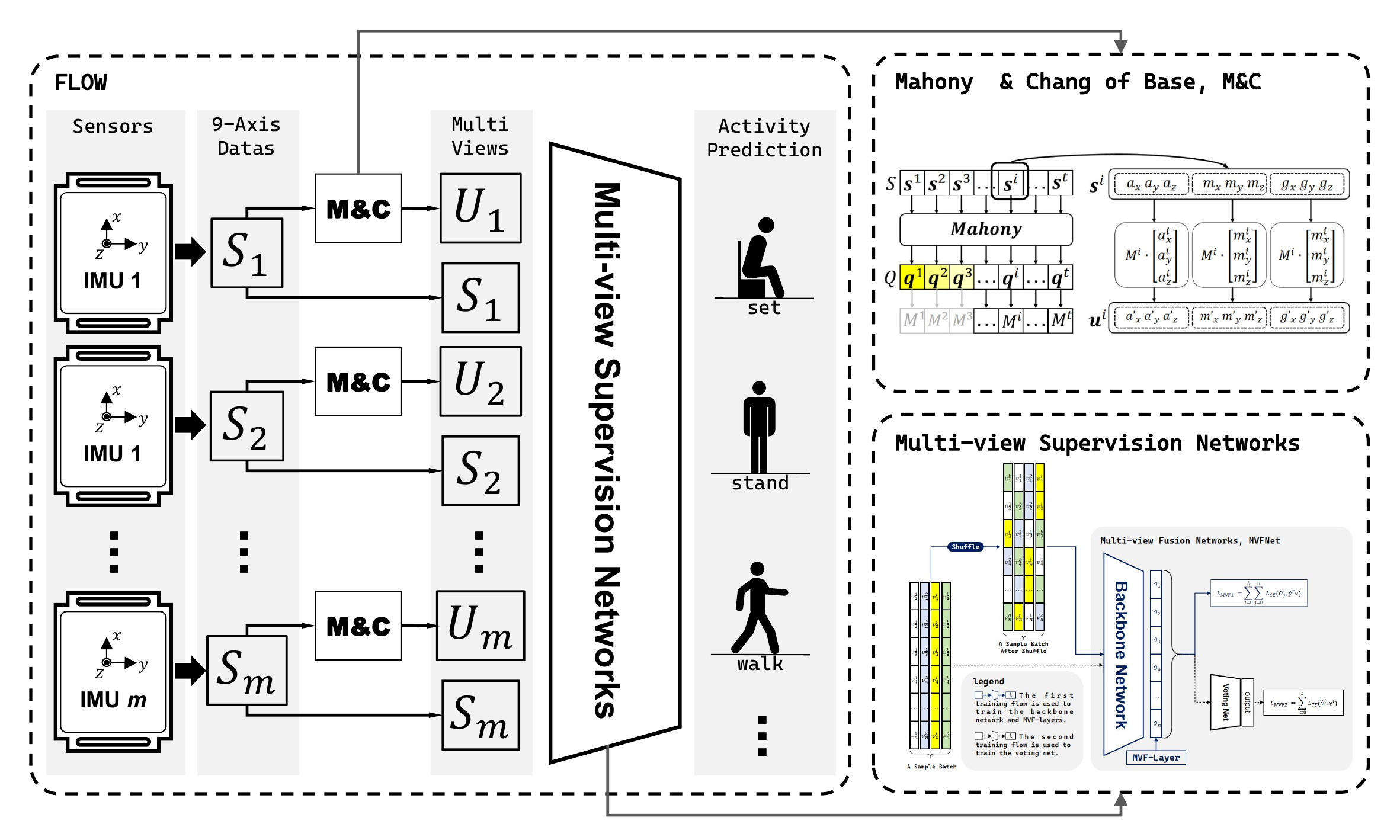}
	\caption{Flow: The M\&C method and MVFNet are combined to obtain a method that can fully utilize the advantages of both local and global views}
	\label{fig4}
\end{figure}

$\boldsymbol{s}_i^j$ represents the raw 9-axis data for sensor $i$ at time $j$, while $\boldsymbol{u}_i^j$ represents the attitude angle data for sensor $i$ at time $j$. Within the FLOW framework, $X$ is treated as a collection of $2m$ views, where each column of $X$ corresponds to an individual view. MVFNet is then employed for multi-view training. The overall process is illustrated in Fig~\ref{fig4}.

\section{Experiment}
\label{experiment}
This paper evaluates the performance of FLOW for IMU-based Human Activity Recognition (HAR) through a series of experiments. Specifically, we conduct the following:

\begin{itemize}
	\item \textbf{Comparative Experiments:} We establish FLOW's superiority on the HAR task by comparing it against established cross-domain approaches like DANN, JAN, SALIENCE, and GILE. Additionally, we perform a horizontal comparison with recent state-of-the-art cross-user activity recognition methods to verify FLOW's effectiveness. To understand the reasons behind FLOW's efficacy, we analyze the confusion matrices of FLOW and the DeepConvLSTM baseline during this experiment.
	\item \textbf{Ablation Experiments:} We delve into the role of each component within FLOW through ablation experiments. Here, we conduct a more in-depth investigation of the impact exerted by MVFNet on the final performance.
\end{itemize}

The following details the full experimental setup.

\subsection{Dataset Introduction}
While the proposed method requires a relatively long data sequence for effective posture feature extraction, pre-segmented datasets like UCI \cite{anguita2013public} are unsuitable due to their processing methods (this limitation is not encountered in practical applications). In the following experiments, we leverage two publicly available datasets that satisfy the experiment's requirements. These datasets are introduced below:

\begin{itemize}
	\item \textbf{PAMAP2:} This dataset encompasses daily activities performed by four participants in a home environment. It involves 18 fine-grained hand gesture activities, such as opening/closing dishwashers, refrigerators, and drawers. Each participant wears 113-dimensional inertial sensors positioned at 12 body locations \cite{reiss2012introducing}.
	\item \textbf{OPPORTUNITY:} This dataset includes data from six participants wearing IMU sensors on their chests, wrists, and ankles. Each IMU sensor is equipped with an accelerometer, gyroscope, and magnetometer. Participants were instructed to perform 12 protocol activities and 6 optional activities \cite{anguita2013public}.
\end{itemize}

The necessary information statistics of the dataset are presented in the Table~\ref{table:dataset}.

\begin{table}[h]
	\centering
	\footnotesize
	\caption{Datasets Information}
	\label{table:dataset}
	\begin{tabular}{ccccc}
		\hline
		\textbf{Datasets} & \textbf{Subject} & \textbf{Rate} & \textbf{Sample} & \textbf{Classes} \\ \hline
		PAMAP2 & 9 & 30 Hz & 2,872,533 & 18 \\ \hline
		OPPO & 4 & 30 Hz & 701,366 & 18 \\ \hline
	\end{tabular}
\end{table}

\subsection{Experiments Setup}

We implemented our model using PyTorch and trained it with the Adam optimizer. Each experiment was trained for 300 epochs. Model evaluation employed two metrics: accuracy ($acc$) and F1-score ($F_1$).

\begin{align}
	acc = \frac{TP}{TP+FN+FP+TN}
\end{align}

where $TP, TN$ is  the number of correctly predicted positive and negative instances, and $FP, FN$ is the number of instances that are predicted as positive but are actually negative and are predicted as negative but are actually positive.

\begin{align}
	F_1 = \sum_{i=1}^k \frac{ 2w_i \times TP_i}{2TP_i + FP_i + FN_i}
\end{align}

where $TP_i$, $FP_i$ and $FN_i$ represent the number of true positives, false positives and false negatives of a class $i$, and and $w_i$ is the fraction of $i^{th}$ classes, respectively. The number of classes is given by $k$. 

The optimal view partitioning schemes for each dataset were determined through ablation experiments (detailed in Section~\ref{sec:gen}). Specifically, OPPORTUNITY dataset utilizes a large granularity partitioning scheme, while PAMAP2 dataset benefits from a medium granularity scheme as Table~\ref{tab:difviews}
 
\subsection{Comparative Experiments}
To demonstrate FLOW's efficacy in cross-user HAR, we compare it against cross-domain methods like DANN, JAN, SALIENCE, and GILE, using DeepConvLSTM as the baseline. The experimental setup employs the "leave-one-out" cross-validation scheme, where the model is trained on the data of n-1 users and tested on the remaining user.

Table~\ref{tab:comparison} presents the test set accuracy results. Most data in the table comes from our own experiments, except for GILE on OPPORTUNITY and SALIENCE on PAMAP2, which are from the respective original papers.

\begin{sidewaystable}
	\centering
	\footnotesize
	\caption{Comparison of FLOW with Cross-Domain Approach}
	\label{tab:comparison}
	\begin{tabularx}{\textwidth}{ c c c c c c c c}
		\cline{1-8}
		\textbf{Datasets}& \textbf{Source to Traget} & \textbf{DeepConvLSTM\cite{ordonez2016deep}} & \textbf{DANN\cite{ganin2016domain}} & \textbf{JAN\cite{long2017deep}} & \textbf{SALIENCE\cite{chen2022salience}} & \textbf{GILE\cite{qian2021latent}} & \textbf{FLOW(ours)}\\
		\cline{1-8}
		\multirow{10}{*}{\textbf{PAMAP2}} & 2,3,4,5,6,7,8,9 to \textbf{1} & 0.7659 & \textbf{0.8059} & 0.7625 & 0.7760 & 0.6720 & 0.7586 \\
		& 1,3,4,5,6,7,8,9 to \textbf{2} & 0.7920 & 0.8867 & 0.9058 & 0.9270 & 0.8564 & \textbf{0.9036}  \\
		& 1,2,4,5,6,7,8,9 to \textbf{3} & 0.9535 & 0.9426 & 0.9406 & 0.8680 & 0.9258 & \textbf{0.9626}  \\
		& 1,2,3,5,6,7,8,9 to \textbf{4} & 0.9385 & 0.9604 & 0.9567 & 0.9470 & 0.9414 & \textbf{0.9728}  \\
		& 1,2,3,4,6,7,8,9 to \textbf{5} & 0.8981 & 0.8870 & 0.8675 & 0.9000 & 0.8792 & \textbf{0.8947} \\
		& 1,2,3,4,5,7,8,9 to \textbf{6} & 0.8917 & 0.9260 & \textbf{0.9296} & 0.9300 & 0.9025 & 0.9116 \\
		& 1,2,3,4,5,6,8,9 to \textbf{7} & 0.9642 & 0.9537 & 0.9540 & 0.9110 & 0.9460 & \textbf{0.9714} \\
		& 1,2,3,4,5,6,7,9 to \textbf{8} & 0.4427 & 0.7360 & 0.6217 & \textbf{0.8920} & 0.4587 & 0.7933 \\
		& 1,2,3,4,5,6,7,8 to \textbf{9} & 0.9752 & 0.4516 & 0.8000 & 0.2483 & 0.6285 & \textbf{1.0000} \\
		\cline{2-8}
		& \textbf{Average} & 0.8469 & 0.8388 & 0.8598 & 0.8222 & 0.8011 & \textbf{0.9076} \\
		\cline{1-8}
		\multirow{5}{*}{\textbf{OPPO}} & 2,3,4 to \textbf{1} & 0.8363 & 0.8386 & 0.8345 & 0.7431 & 0.8386 & \textbf{0.8815} \\
		& 1,3,4 to \textbf{2} & 0.8347 & 0.8401 & 0.8345 & 0.7996 & 0.8165 & \textbf{0.8387} \\
		& 1,2,4 to \textbf{3} & 0.7659 & 0.7757 & 0.7776 & 0.7213 & 0.7866 & \textbf{0.7905} \\
		& 1,2,3 to \textbf{4} & 0.8026 & 0.8109 & 0.8129 & 0.7790 & 0.8141 & \textbf{0.9021} \\
		\cline{2-8}
		& \textbf{Average} & 0.8098 & 0.8163 & 0.8148 & 0.7608 & 0.8140 & \textbf{0.8532} \\
		\cline{1-8}
	\end{tabularx}
\end{sidewaystable}

Table~\ref{tab:comparison} demonstrates FLOW's superiority over existing mainstream cross-domain solutions on both PAMAP2 and OPPORTUNITY datasets. Compared to DeepConvLSTM, which lacks any cross-domain adaptation, FLOW achieves significant improvements. Specifically, FLOW reduces the error rate by 39.21\% and 23.03\% on PAMAP2 and OPPORTUNITY, respectively.

Furthermore, we benchmark FLOW against the latest state-of-the-art methods reported in the literature, using evaluation metrics from their respective original papers. As shown in Table~\ref{tab:comparison2}, FLOW outperforms all other approaches across all metrics.

\begin{table}
	\caption{Comparison of FLOW with Other SOAT Approach}
	\label{tab:comparison2}
	\centering
	\begin{tabular}{ c c c| c| c| c| c c} 
		\hline
		\multirow{2}{*}{} & \multicolumn{2}{c}{\textbf{TASKED\cite{suh2023tasked}}} & \textbf{CNN-LSTM+\cite{tang2022dual}} & \textbf{InnoHAR+\cite{tang2022dual}} & \textbf{DIN\cite{leite2020improving}} & \multicolumn{2}{c}{\textbf{FLOW}(ours)}  \\ 
		\cline{2-8}
		& $acc$ & $F_1$ & $F_1$ & $F_1$ & $F_1$ & $acc$ & $F_1$ \\ 
		\hline
		\textbf{PAMAP2} & 0.8304 & 0.8293 & 0.7259 & 0.8130 & 0.9044 & \textbf{0.9076} & \textbf{0.9056} \\ 
		\hline
		\textbf{OPPO} 	& 0.7583 & 0.7583 & 0.6318 & 0.6765 & / & \textbf{0.8532} & \textbf{0.8527} \\
		\hline
	\end{tabular}
\end{table}

\subsection{Ablation Experiments}
This experiment investigates the individual contributions of the M\&C part and the multi-view fusion part (MVFNet) within FLOW, and analyzes their training details through ablation experiments.

The ablation experiments are divided into two groups:

\begin{itemize}
	\item \textbf{Group 1: Impact of Different Views: }This group isolates the influence of different views on the model's cross-user performance by excluding MVFNet. We employ the DeepConvLSTM model as a classifier to categorize two separate representations, $\boldsymbol{v}_L$, $\boldsymbol{v}_G$, and $\boldsymbol{v}_L + \boldsymbol{v}_G$, without multi-view fusion. This analysis aims to assess the impact of these representations on the model's generalization ability.
	\item \textbf{Group 2: View Granularity in FLOW: }This group explores the influence of view granularity selection within the FLOW model. We will experiment with four different view division methods on two datasets to evaluate the impact of view division on the final results.
\end{itemize}

\subsubsection{Effect of Different Generalization of Views} \label{sec:gen}
To analyze the impact of different viewpoints on the model's generalization performance, we employ the DeepConvLSTM model as the classifier and feed it with the representations  $\boldsymbol{v}_L$, $\boldsymbol{v}_G$ and $\boldsymbol{v}_L + \boldsymbol{v}_G$ as input, respectively. Right of Table~\ref{tab:difviews} presents the results of this first ablation experiment.

\begin{table}
	\footnotesize
	\centering
	\caption{The effect of different views and view division granularity}
	\label{tab:difviews}
	\begin{tabular}{ c c c c c| c c c c}
		\hline
		\multirow{2}{*}{\textbf{Datasets}} & \multirow{2}{*}{\textbf{Source to Traget}} & \multicolumn{3}{c}{\textbf{DeepConvLSTM(Views)}} & \multicolumn{4}{c}{\textbf{FLOW(Views Division)}} \\
		\cline{3-9}
		& & $\boldsymbol{v}_L$ & $\boldsymbol{v}_G$ & $\boldsymbol{v}_L+\boldsymbol{v}_G$ & \textbf{$\boldsymbol{v}_L$} & \textbf{small} & \textbf{medium} & \textbf{large}\\
		\cline{1-9}
		\multirow{10}{*}{\textbf{PAMAP2}} & 2,3,4,5,6,7,8,9 to \textbf{1} & \textbf{0.7659} & 0.7415 & 0.7503 & 0.7280 & 0.7429 & \textbf{0.7586} & 0.7468\\
		& 1,3,4,5,6,7,8,9 to \textbf{2} & 0.7920 & \textbf{0.9092} & 0.8163 & 0.9032 & 0.8429 & 0.9036 & \textbf{0.9138} \\
		& 1,2,4,5,6,7,8,9 to \textbf{3} & 0.9535 & 0.9445 & \textbf{0.9576} & 0.9518 & 0.9564 & 0.9626 & \textbf{0.9664} \\
		& 1,2,3,5,6,7,8,9 to \textbf{4} & \textbf{0.9385} & 0.9313 & 0.9276 & 0.9651 & 0.9589 & \textbf{0.9728} & 0.9535 \\
		& 1,2,3,4,6,7,8,9 to \textbf{5} & \textbf{0.8981} & 0.7297 & 0.8402 & \textbf{0.9015} & 0.8837 & 0.8943 & 0.8309 \\
		& 1,2,3,4,5,7,8,9 to \textbf{6} & \textbf{0.8917} & 0.8594 & 0.8949 & 0.8968 & 0.8976 & \textbf{0.9116} & 0.9063 \\
		& 1,2,3,4,5,6,8,9 to \textbf{7} & \textbf{0.9642} & 0.9255 & 0.9677 & 0.9720 & 0.9698 & 0.9714 & \textbf{0.9723} \\
		& 1,2,3,4,5,6,7,9 to \textbf{8} & 0.4427 & \textbf{0.8022} & 0.5614 & 0.6064 & 0.7863 & \textbf{0.7933} & 0.7627 \\
		& 1,2,3,4,5,6,7,8 to \textbf{9} & 0.9752 & \textbf{1.0000} & 0.9917 & 0.9835 & 0.9834 & \textbf{1.0000} & 0.9834 \\
		\cline{2-9}
		& \textbf{Average} & 0.8469 & \textbf{0.8715} & 0.8564 & 0.8796 & 0.8913 & \textbf{0.9076} &  0.8929 \\
		\cline{1-9}
		\multirow{5}{*}{\textbf{OPPO}} & 2,3,4 to \textbf{1} & 0.8363 & \textbf{0.8742} & 0.8733 & 0.8373 & 0.8090 & 0.8377 & \textbf{0.8815} \\
		& 1,3,4 to \textbf{2} & \textbf{0.8347} & 0.8106 & 0.8055 & 0.8358 & 0.7923 & 0.8380 & \textbf{0.8387} \\
		& 1,2,4 to \textbf{3} & 0.7659 & \textbf{0.8287} & 0.8213 & 0.7910 & 0.7660 & \textbf{0.7963} & 0.7905 \\
		& 1,2,3 to \textbf{4} & 0.8026 & 0.8900 & \textbf{0.9061} & 0.8645 & 0.8084 & 0.8625 & \textbf{0.9021} \\
		\cline{2-9}
		& \textbf{Average} & 0.8098 & 0.8508 & \textbf{0.8515} & 0.8325 & 0.7918 & 0.8336 & \textbf{0.8532}\\
		\cline{1-9}
	\end{tabular}
\end{table}

Table~\ref{tab:difviews} reveals that $\boldsymbol{v}_G$ on average outperforms $\boldsymbol{v}_L$ on both datasets. However, this is not always the case in each row of the comparison, possibly due to information loss during the conversion process to $\boldsymbol{v}_G$.


Compared to $\boldsymbol{v}_L$, $\boldsymbol{v}_G$ exhibits more stable generalization performance. For instance, on the PAMAP2 dataset, the model using $\boldsymbol{v}_L$ as input performs poorly on cross-user tasks when user-2 and user-8 are used for testing. This is not observed with $\boldsymbol{v}_G$.

Our experiments demonstrate that simply concatenating $\boldsymbol{v}_G$ and $\boldsymbol{v}_L$ does not guarantee optimal results. In most cases, the performance falls between that of $\boldsymbol{v}_G$ and $\boldsymbol{v}_L$. We hypothesize that this occurs because the model overlooks crucial information, which is precisely why MVFNet is employed.

To delve deeper into the impact of $\boldsymbol{v}_G$ and $\boldsymbol{v}_L$ on specific categories, we analyze the confusion matrix for a representative experimental group. The experiment on the PAMAP2 dataset with user-8 as the test set is chosen for this analysis.

\begin{figure}[htbp]
	\label{img:mtr}
	\centering
	\begin{subfigure}[b]{0.45\textwidth} 
		\includegraphics[width=\textwidth]{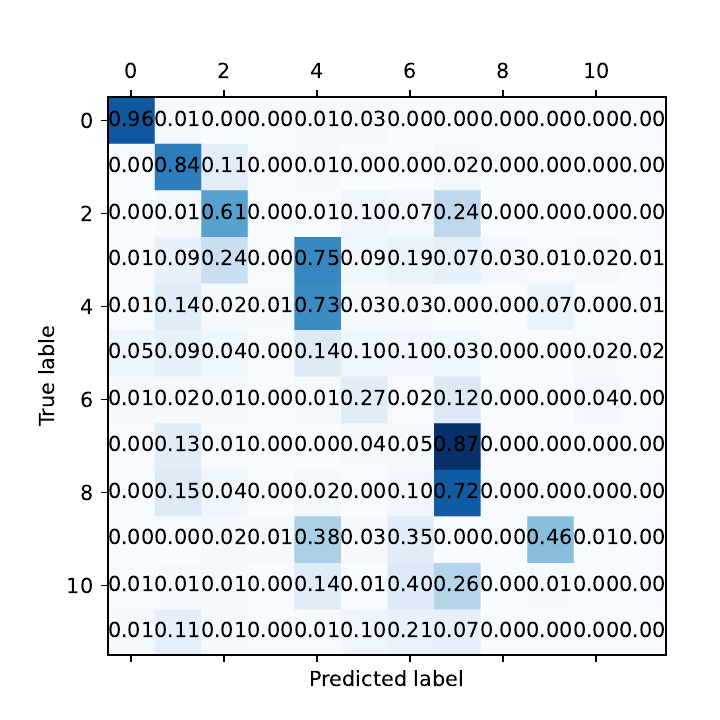} 
		\caption{DeepConvLSTM traing by $\boldsymbol{v}_L$}
		\label{fig:a}
	\end{subfigure}
	\hfill
	\begin{subfigure}[b]{0.45\textwidth} 
		\includegraphics[width=\textwidth]{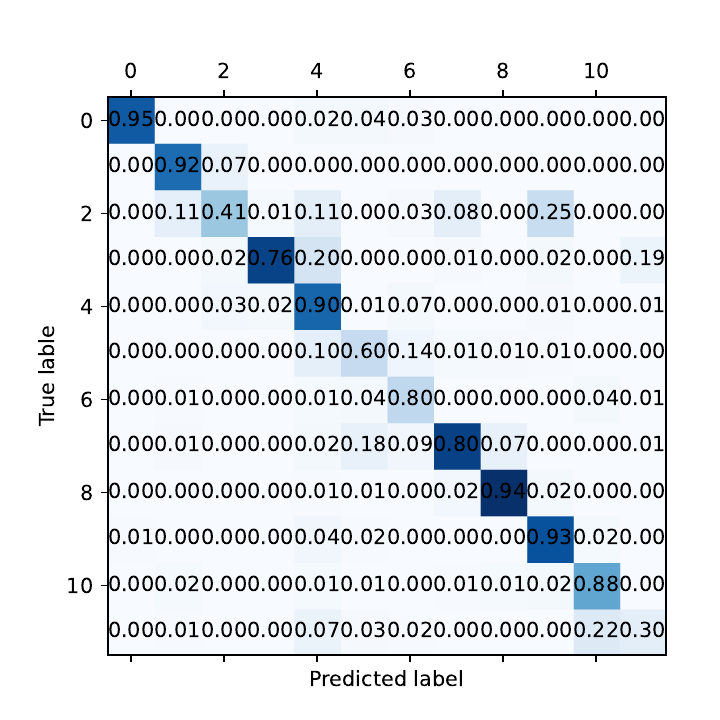} 
		\caption{DeepConvLSTM traing by $\boldsymbol{v}_G$}
		\label{fig:b}
	\end{subfigure}
	\caption{Confusion matrix obtained by training with different views}
	\label{fig:both}
\end{figure}

Figure~\ref{fig:a} presents the confusion matrix for the model when using $\boldsymbol{v}_L$ as input, Figure~\ref{fig:b} shows the corresponding matrix for $\boldsymbol{v}_G$. Both inputs achieve good performance for basic actions (e.g., 1. lying, 2. sitting, 3. standing). However, for complex actions (e.g., 9. vacuuming a room), $\boldsymbol{v}_G$ demonstrates a clear advantage over $\boldsymbol{v}_L$.

\subsubsection{View Granularity Analysis in MVFNet}
The granularity of view division plays a crucial role in influencing model performance. On one hand, larger view divisions can provide richer information for each view, potentially enhancing the performance of each classifier within the MVF-layer of MVFNet. On the other hand, smaller view divisions lead to a higher number of classifiers, which can supply the voting layer with more results for comparison and selection.

To quantitatively investigate the impact of view division granularity on model performance, this subsection employs three view division schemes for experiments:

\begin{enumerate}
	\item \textbf{Just $\boldsymbol{v}_L$: } Only $\boldsymbol{v}_L$ data is used, and the data of each IMU is regarded as a view, that is, if there are $m$ IMUs, $m$ views will be divided. It is used to study the performance of MVFNet separately with the influence of $\boldsymbol{v}_G$ removed.
	\item \textbf{Small Granularity Scheme:} Each set of measurement values of each sensor is considered as a view. For example, a sensor can be divided into 7 views: $[a_x, a_y, a_z], [ m_x, m_y, m_z]$, $ [g_x, g_y, g_z], [a'_x, a'_y, a'_z], [ m'_x, m'_y, m'_z], [g'_x, g'_y, g'_z], [q_x, q_y, q_z, q_w]$. If there are $m$ sensors, then there are $7m$ views.
	
	\item \textbf{Medium Granularity Scheme:} Each sensor is divided into two views: $\boldsymbol{v}_L$ and $\boldsymbol{v}_G$. If there are $m$ sensors, then there are $2m$ views.
	
	\item \textbf{Large Granularity Scheme:} The $\boldsymbol{v}_L$ and $\boldsymbol{v}_G$ of all sensors are considered as two views together. The number of views in this scheme is not related to the number of sensors.
\end{enumerate}

We evaluate model performance using accuracy, and the results are presented in Table~\ref{tab:difviews}.

Table~\ref{tab:difviews} demonstrates that FLOW($\boldsymbol{v}_L$) has a good improvement compared to DeepConvLSTM($\boldsymbol{v}_L$), indicating that even when using only a single view, MVFNet can still improve the cross-user performance of IMU-HAR.

\begin{figure}[htbp]
	
	\centering
	\begin{subfigure}[b]{0.45\textwidth} 
		\includegraphics[width=\textwidth]{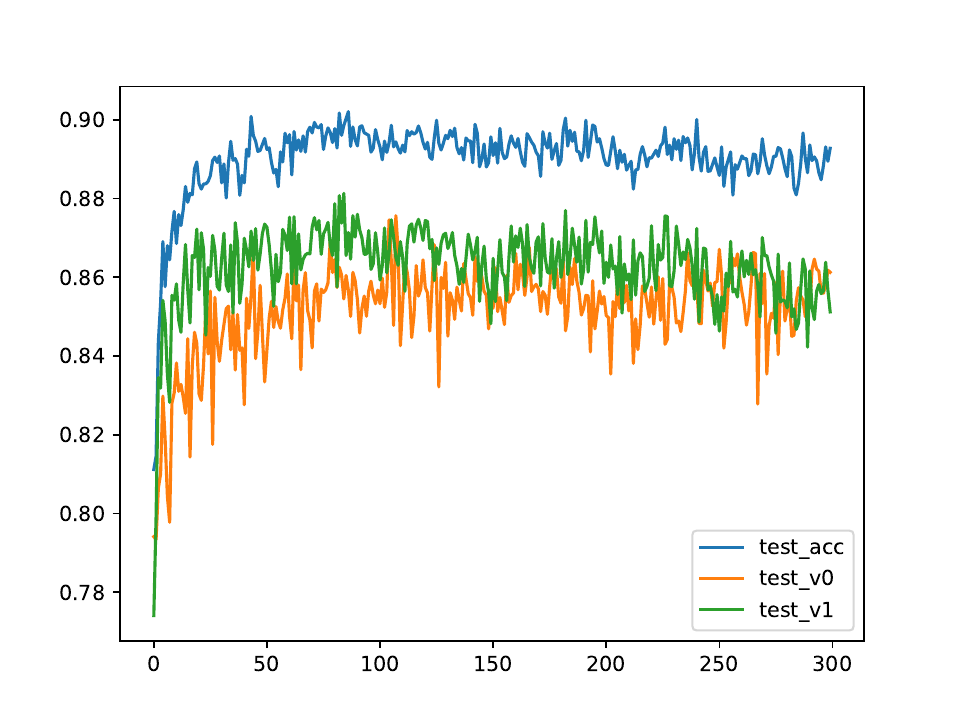} 
		\caption{Training process at large granularity scheme in OPPO}
		\label{fig:big}
	\end{subfigure}
	\hfill
	\begin{subfigure}[b]{0.45\textwidth} 
		\includegraphics[width=\textwidth]{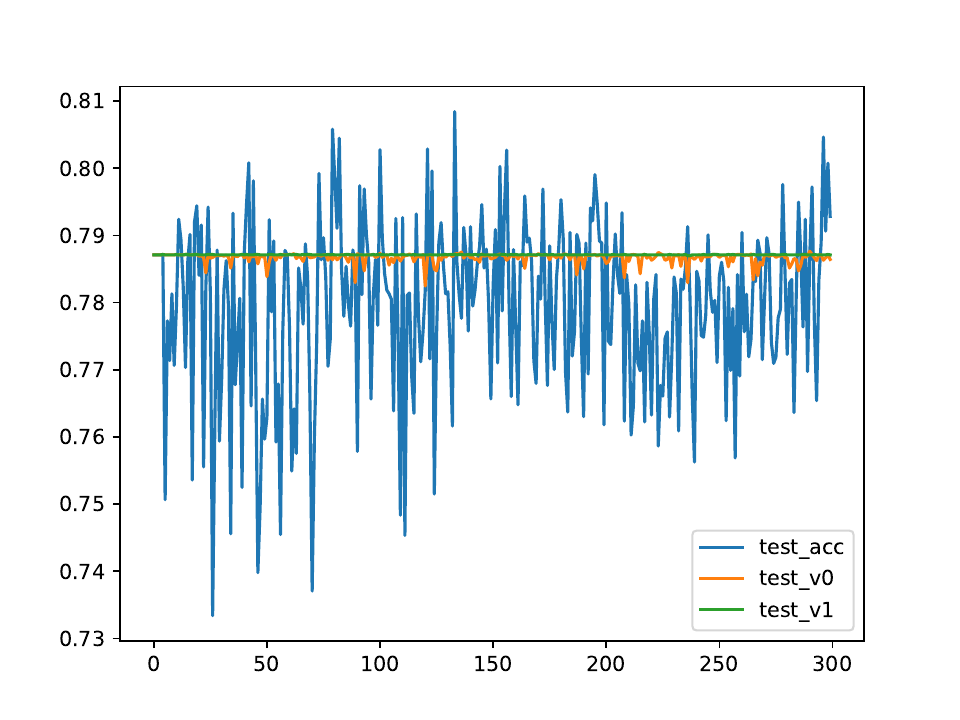} 
		\caption{Training process at small granularity scheme in OPPO}
		\label{fig:small}
	\end{subfigure}
	\caption{Training process under different view granularity}
	\label{img:zxt}
\end{figure}

The experimental results confirm the influence of view granularity on model performance. Notably, on the OPPORTUNITY dataset, which exhibits a stronger sensitivity, the Large group scheme achieves significantly better results compared to the Small group scheme. Figure~\ref{img:zxt} illustrates the accuracy curve of the model training process when using user-4 as the test set.

Fig~\ref{fig:big} and~\ref{fig:small} illustrate the results for the large and small view group experiments, respectively. To maintain clarity, each figure only presents the final output's test accuracy (test\_acc) and the test accuracies (test\_v0 and test\_v1) for the classifiers corresponding to the first two views.

The large group's accuracy reaches a peak of approximately 90\% after 100 iterations, followed by a gradual decline due to overfitting. In contrast, the small group's accuracy oscillates between 81\% and 73\% after reaching 78\% on the test set. It exhibits minimal improvement with increasing iterations, suggesting that the model converges quickly, potentially reaching its optimal level after the first iteration.

Based on the figures, we hypothesize that the small group's lower accuracy stems from insufficient information for each sub-classifier in the MVF-layer to make accurate judgments. As the adage goes, 'aggregating garbage sub-classifiers can only get a garbage general classifier'. Consequently, the voting network cannot be expected to produce a high-quality final output.

Conversely, the large group achieves a higher final output quality despite having only two classifiers in the MVF-layer, due to the strong performance of both individual classifiers.

It is important to note that all FLOW experiments presented earlier in the paper employed the segmentation granularity most suitable for each dataset, as determined by Table~\ref{tab:difviews}. Specifically, medium granularity segmentation was used for the PAMAP2 dataset, and large granularity segmentation was used for the OPPORTUNITY dataset.

\subsection{Experiment Summary}

This paper presents a comprehensive investigation of FLOW, a novel method for cross-user Human Activity Recognition (HAR) using Inertial Measurement Unit (IMU) data. The evaluation involves two key parts:

\textbf{Part 1: Benchmarking FLOW against Existing Methods}

We compared FLOW with four established cross-domain methods and three recent state-of-the-art cross-user methods on the publicly available PAMAP2 and OPPORTUNITY datasets. The results demonstrate that FLOW outperforms all the compared methods, establishing its superiority in cross-user HAR tasks.

\textbf{Part 2: Ablation Experiments and Analysis}

The second part of the study focused on ablation experiments to gain deeper insights into FLOW's internal workings. We conducted these experiments on public datasets and verified that the globally encoded representation, $\boldsymbol{v}_G$, is a more robust data representation scheme compared to the locally encoded representation, $\boldsymbol{v}_L$, for tackling cross-user problems. Notably, $\boldsymbol{v}_G$ exhibits a significant advantage in recognizing complex activities. Additionally, we investigated the influence of view granularity within the multi-view fusion component. The results suggest that excessively small view partitions should be avoided, as they can limit the performance of individual sub-classifiers due to data insufficiency.

\textbf{Key Findings:}

\begin{enumerate}
	\item \textbf{Superior Cross-User Performance:} FLOW surpasses existing benchmarks for cross-user HAR tasks.
	\item \textbf{Robust Representation for Cross-User Scenarios:} The $\boldsymbol{v}_G$  offers a more stable and effective solution for cross-user problems compared to the $\boldsymbol{v}_L$. This is particularly evident in recognizing complex activities.
	\item \textbf{Optimal View Granularity: } The granularity of view partitioning within FLOW should be carefully chosen. Partitions that are too small can hinder the performance of sub-classifiers due to limited data, resulting in a decrease in overall model accuracy.
\end{enumerate}

\section{Conclusion}
\label{conclusion}
This paper presents a novel Inertial Measurement Unit-based Human Activity Recognition (IMU-HAR) method that leverages basis transformation and multi-view fusion. The core concept involves first extracting $\boldsymbol{v}_G$ of the IMU data $\boldsymbol{v}_L$ using an attitude solution algorithm and a basis transformation operation within the NED coordinate system. Subsequently, a Multi-View Fusion Network (MVFNet) is employed to effectively fusion $\boldsymbol{v}_L$ and $\boldsymbol{v}_G$, to make sure the model doesn't ignore critical information.

The proposed method achieves superior results compared to several established cross-domain methods and current state-of-the-art approaches on benchmark datasets. These findings highlight the potential of data-centric methods for tackling cross-user HAR challenges.

One limitation of the proposed method is the additional computational cost associated with the explicit attitude solution step. Future research directions include incorporating deep learning techniques to enable the model to automatically learn and extract diverse feature views, eliminating the need for manual design. Additionally, we aim to explore the applicability of MVFNet in other research domains to further evaluate its capabilities.


\end{document}